%% file: root.tex
\documentclass[letterpaper, 10 pt, conference]{ieeeconf}  

\IEEEoverridecommandlockouts                              

\usepackage{comment}

\usepackage{glossaries}
\input{ext/abb}

\input{ext/symbols}

\usepackage{amsmath}
\usepackage{amsfonts}

\usepackage{siunitx}

\usepackage{graphicx}
\usepackage{multirow}
\usepackage{hhline}
\usepackage{adjustbox}
\usepackage{tabularray}

\usepackage{subcaption}

\usepackage[backend=biber,style=ieee,mincitenames=1,maxcitenames=2,natbib=true]{biblatex}
\usepackage{doi}

\usepackage{xcolor}

\newcommand{\paraDraft}[1]{\ifdefined\draft\subsubsection*{\color{blue}\textbf{#1}}\fi}

\providecommand{\ie}{\textit{i.e.,}~} %
\providecommand{\sectionname}{Section}
\providecommand*{\sref}[1]{\sectionname~\ref{s:#1}}            
\providecommand{\tablename}{Table}
\providecommand*{\tref}[1]{\tablename~\ref{t:#1}}   
\providecommand{\figurename}{Fig.}
\providecommand*{\fref}[1]{\figurename~\ref{f:#1}}  
\providecommand*{\fref}[1]{\figurename~\ref{f:#1}}  
\providecommand{\equationname}{Eq.}
\providecommand*{\eref}[1]{\equationname~(\ref{e:#1})}            

\title{\LARGE \bf
Few-Shot Learning of Force-Based Motions From Demonstration \\ Through Pre-training of Haptic Representation
}

\author{Marina Y. Aoyama$^{1}$, Jo\~{a}o Moura $^{1,2}$, Namiko Saito$^{1,2}$ and Sethu Vijayakumar$^{1,2}$
\thanks{$^{1}$Authors are with the School of Informatics, The University of Edinburgh, Edinburgh, U.K.}
\thanks{$^{2}$Authors are with The Alan Turing Institute, London, U.K.}%
}

\begin{document}

\maketitle
\thispagestyle{empty}
\pagestyle{empty}

\begin{abstract}
In many contact-rich tasks, force sensing plays an essential role in adapting the motion to the physical properties of the manipulated object. 
To enable robots to capture the underlying distribution of object properties necessary for generalising learnt manipulation tasks to unseen objects, existing \gls{lfd} approaches require a large number of costly human demonstrations.
Our proposed semi-supervised \gls{lfd} approach decouples the learnt model into an haptic representation encoder and a motion generation decoder.
This enables us to pre-train the first using large amount of unsupervised data, easily accessible, while using few-shot \gls{lfd} to train the second, leveraging the benefits of learning skills from humans.
We validate the approach on the wiping task using sponges with different stiffness and surface friction. 
Our results demonstrate that pre-training significantly improves the ability of the \gls{lfd} model to recognise physical properties and generate desired wiping motions for unseen sponges, outperforming the \gls{lfd} method without pre-training. 
We validate the motion generated by our semi-supervised \gls{lfd} model on the physical robot hardware using the KUKA iiwa robot arm.
We also validate that the haptic representation encoder, pre-trained in simulation, captures the properties of real objects, explaining its contribution to improving the generalisation of the downstream task.  
\end{abstract}

\section{INTRODUCTION}
\label{s:introduction}
\paraDraft{Motivate need of adaptation of motion based on object's physical properties}

Humans are able to adapt their motion according to the different physical properties of the objects they manipulate by extensively using force and tactile sensing~\cite{Valero2017_human_haptic_manipulation, Dahiya2009_human_tactile, Lederman1987_exploratory_procedure_manipulation}.
For instance, when wiping a surface with a sponge, humans naturally adapt their wiping motion based on the stiffness and friction of both the table and sponge, in order to maximise cleaning effectiveness while avoiding damage~\cite{kawaharazuka2022_humanoid_wiping_lfd, Do_ICRA2014_adaptive_wiping}. 
Similarly, in various other robotic contact-rich manipulation tasks such as food manipulation~\cite{gemici2014_iros_saxena, sundaresan2022_corl_skewering, Bhattacharjee2019_RA-L_haptic_assistive_feeding_food}, pouring~\cite{LopezGuevara2020_iros_tatiana, Saito2019_robio_namiko_pouring}, or peg-in-hole insertion~\cite{Davchev2022_RA-L_todor_residualLfD_insertion}, the ability to adapt the motion according to the physical properties of the manipulated objects is vital for the successful execution of the task.

\paraDraft{Challenge learning manipulation tasks generalisable across objects}
\glsfirst{lfd} is an intuitive and efficient approach for robots to acquire skills or behaviours by imitating human demonstrations~\cite{Billard2008_lfdsurvey_efficient_intuitive, Hussein2017_lfdsurvey_humanskill, Ravichandar2020_lfd_survey}. 
One of the challenges in~\gls{lfd} is how to generalise learnt manipulation motions to objects with different physical properties. 
In addition, producing human demonstrations requires some cognitive and physical effort, typically resulting in few examples with a small variety of objects~\cite{Wang2021_ACM_few-shot_survey, James2018_CoRL_few-shotLfD_vision_RL}. 
Therefore, we address the question -- how can robots learn to manipulate unseen objects with different physical properties from a small number of human demonstrations? 

\begin{figure}[t]
\centering
\includegraphics[keepaspectratio, scale=0.3]{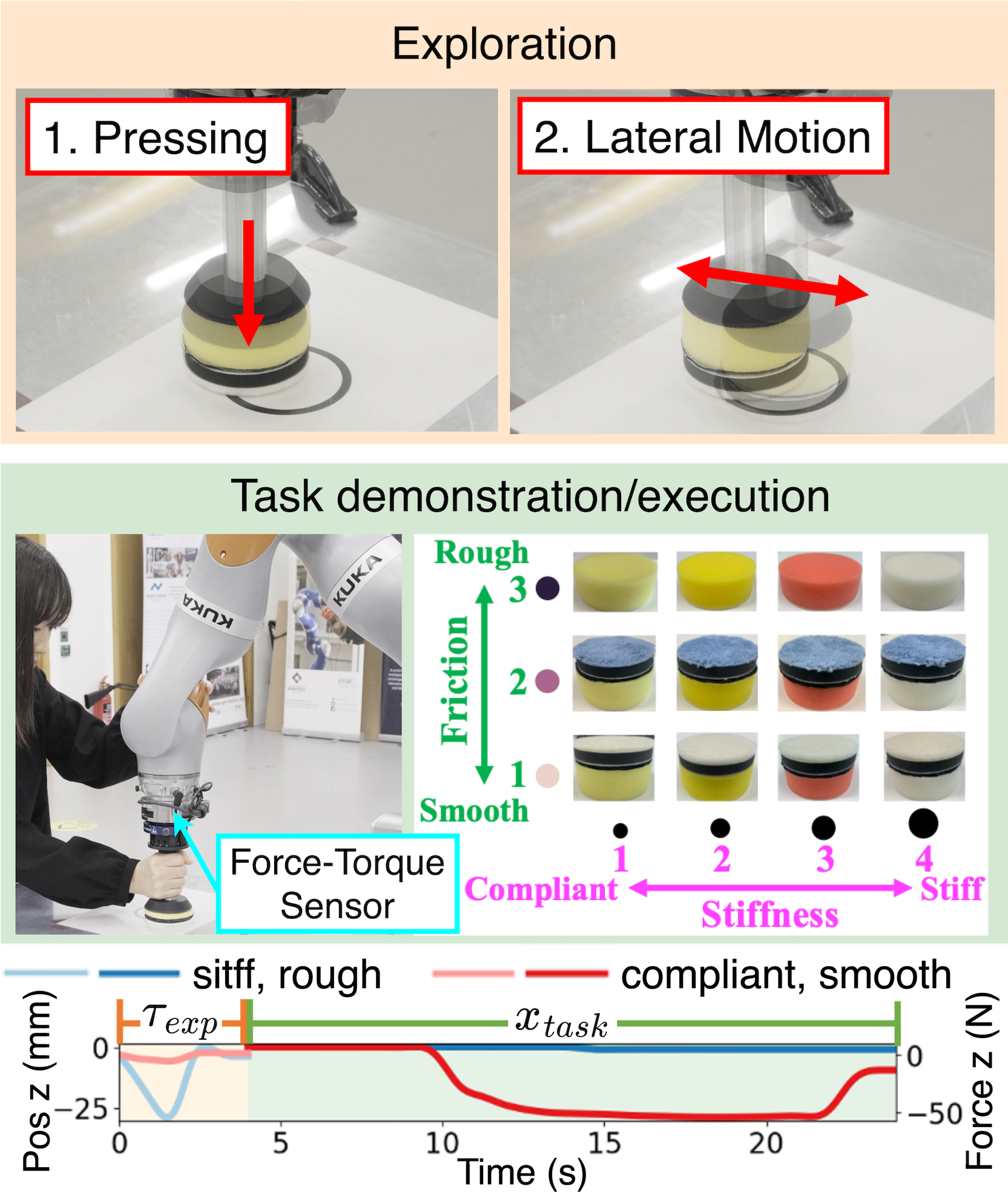}
\caption{
Wiping Experiment. 
In the exploration phase, the robot performs exploratory actions to identify the physical properties of the sponge. 
Then, in the task execution phase, the robot performs the desired wiping motion according to the physical properties of the sponge identified in the exploration phase.
}
\setlength\intextsep{0pt}
\label{f:experiment_setup}
\vspace{-0.5cm}
\end{figure}

\subsection{Related Work}
There are previous works in~\gls{lfd} that consider physical properties of objects in order to adapt the learnt motion~\cite{sundaresan2022_corl_skewering, Saito2019_robio_namiko_pouring, saito2021_ral_namiko_bestpaper}. 
These works achieve adaptation by performing exploratory actions on the manipulated object to obtain the latent state of the learnt target task, with the assumption that that latent state implicitly encodes the object characteristics.
However, there is a coupling between the exploration and target tasks in the sense that the robot needs to jointly learn both the exploration and the task models, the first responsible for characteristic encoding and the second for motion generation.
Therefore, in order to capture the underlying distribution of object properties and improve generalisation, one needs to increase the number of demonstrations, incurring a significant amount of human effort.

\paraDraft{Pre-training and semi-supervised learning}
In the context of representation learning, there are multiple studies addressing the pre-training of perceptual models using unsupervised data~\cite{Merckling2020_book_representation_learning_lfd, bengio2013_TPAMI_representation_semi-supervised}. 
Particularly in robotics, pre-training visual representations leads to improved robustness of the learnt tasks to different image observations~\cite{pari2021_rss_visual_pretrain, Florence2019_RA-L_self-supervised_pre-training_lfd_vision, radosavovic2022_CoRL_pretraining_vision}. 
While visual representation learning benefits from widely available image datasets~\cite{Salari2022_neurocomputing_visual_imagedataset} or pre-trained models~\cite{HAN2021_aiopen_visual_pretrainedmodel_survey}, what remains unclear is how to acquire large-scale unsupervised data that effectively captures the physical properties of the objects and how to pre-train models that encode these properties using force information. 

Force or tactile perception can reveal physical properties of objects, such as weight, texture, and compliance that are typically unavailable through visual perception~\cite{Bohg2017_Trans_interactive_perception, natale2004_physical_interactive_perception, Chu2015_robot_interactive_perception_haptic}. 
To address this,~\citet{guzey2023_arxiv_tactile_pretrain} pre-train a representation based on tactile observations.
However, none of these visual~\cite{pari2021_rss_visual_pretrain,Florence2019_RA-L_self-supervised_pre-training_lfd_vision} or tactile~\cite{guzey2023_arxiv_tactile_pretrain} approaches applies self-supervised learning on the time series of observations.
In contrast, there are works that highlight the importance of force or tactile time-series, obtained through interactive exploration, in effectively capturing physical object properties~\cite{Drimus2014_time_series_deformable_danicakragic, Chu2015_robot_interactive_perception_haptic, Kroemer2011_tactile_timeseries}, albeit without ever using that information for motion adaptation.
Therefore, we propose an haptic representation that considers a series of force observations and address the question of how to adapt motion based on the obtained object characteristics.



\subsection{Problem Formulation}
\label{s:problem}

\paraDraft{Problem setting}
Similarly to the prior works on adapting motion based on the physical properties of the object~\cite{Saito2019_robio_namiko_pouring, LopezGuevara2020_iros_tatiana, saito2021_ral_namiko_bestpaper}, we consider a task with two phases: the exploration phase, where the robot performs pre-defined exploratory actions to identify the properties of the object, and the task execution phase, where the robot performs the target task according to the properties of the object identified in the exploration phase, as shown in \fref{experiment_setup}. 
We assume that the environment and the properties of the object remain unchanged during the task.

\paraDraft{Objective}
The objective of \gls{lfd} is to enable robots to perform manipulation tasks similar to human demonstrators. 
Therefore, we aim to learn a model $\lfddecoder$ that, given the force trajectory $\texp$ obtained through the exploration, generates a motion trajectory $\xtask$ for the target task, similar to the human demonstration. 

\subsection{Contribution}
To address the issue of the existing \gls{lfd} approaches that suffer from limited generalisation with few demonstrations, we propose a semi-supervised \gls{lfd} approach that decouples the haptic representation encoder -- the model representing the physical properties of an object -- from the motion generation decoder, which maps the encoded object properties to the desired behaviour.
Through pre-training of the haptic representation encoder enabled by this decoupling, the aim is to utilise large amounts of unsupervised haptic data for generalising the task to unseen objects with different physical properties, even for limited number of demonstrations, while leveraging the benefits of learning skills from humans through \gls{lfd}.

\paraDraft{Problem setting (Data)}
The training data then consists of two types of data sets: unsupervised haptic data and demonstration data.
The unsupervised haptic data consists of $M$ force trajectories $\texp$ for objects with different properties, autonomously collected by repeatedly performing pre-defined exploratory actions either in simulation or on a real robot.
Note that we exclude the labels of the properties from the data set.
The demonstration data consists of $N$ pairs of trajectories, one for each object, composed of a force trajectory $\texp$, autonomously collected as above, and the corresponding motion trajectory $\xtask$, demonstrated by an expert demonstrator to perform the target task. 
$M$ is much larger than $N$ -- the number of available human demonstrations. 
 
\paraDraft{Experiment}
We validate the proposed semi-supervised \gls{lfd} approach on a wiping task using sponges with different stiffness and surface friction.
The task serves as an exemplar force-based manipulation task, as the desired wiping motion depends on the physical properties of the sponge perceived only through observing the interaction force.
Our results suggest that being able to pre-train the haptic representation encoder improves the learning of the wiping motions for unseen sponges.
Moreover, replaying these learnt motions on the physical robot results in force interactions closer to those demonstrated by humans.
Furthermore, a~\gls{tsne} analysis of the model pre-trained in simulation supports the assumption that its latent space indeed captures the object characteristics.

\paraDraft{Contributions}
In summary, the contributions of this work are:
\begin{itemize}
    \item To the best of our knowledge, we propose the first semi-supervised \gls{lfd} framework for pre-training of physical properties from time-series of force information.
    \item We propose a decoupling of the haptic representation encoder from the motion generation decoder, enabling the use of large amounts of unsupervised data to improve the generalisation capability of few-shot \gls{lfd} for objects unseen during demonstration.
    \item We validate that the latent space of the haptic representation encoder, pre-trained in simulation, is able to distinguish the physical properties of the real objects, supporting the assumption that the latent space of our model indeed captures the object characteristics. 
    \item We validate our semi-supervised \gls{lfd} approach on a physical robotic setup, showing that it generates motion that achieves a force profile close to those demonstrated by the human.
\end{itemize}

\renewcommand\thesubfigure{\alph{subfigure}}
\begin{figure*}[t]
\captionsetup[subfigure]{justification=centering}
    \begin{tabular}{ccc}
      \begin{minipage}[b]{0.3\hsize}
        \centering
        \includegraphics[keepaspectratio, scale=0.7]{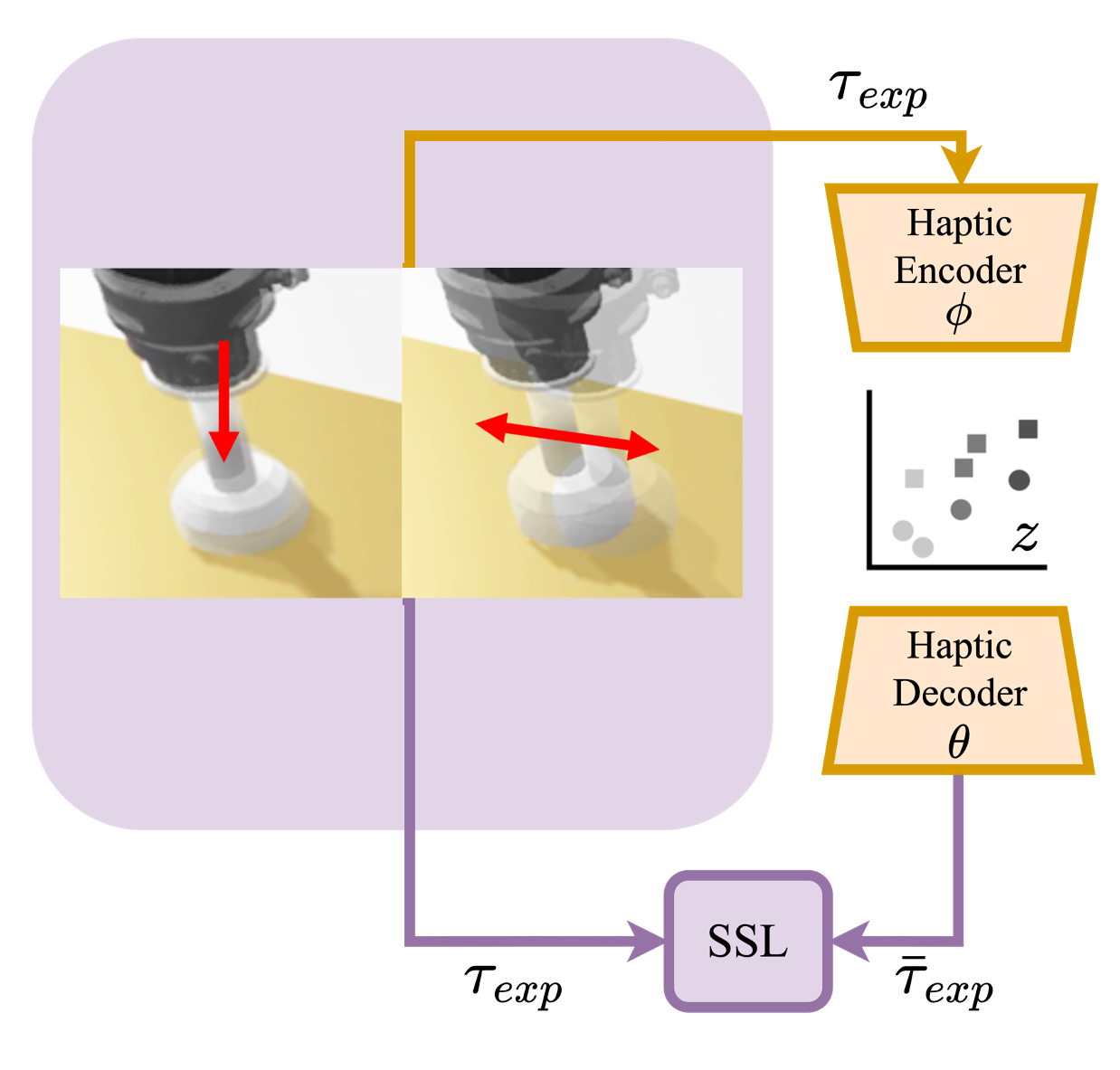}
        \subcaption{Training Step1: \\ Pre-training haptic representation}
        \label{f:training_step1}
      \end{minipage} &
      \begin{minipage}[b]{0.32\hsize}
        \centering
        \includegraphics[keepaspectratio, scale=0.7]{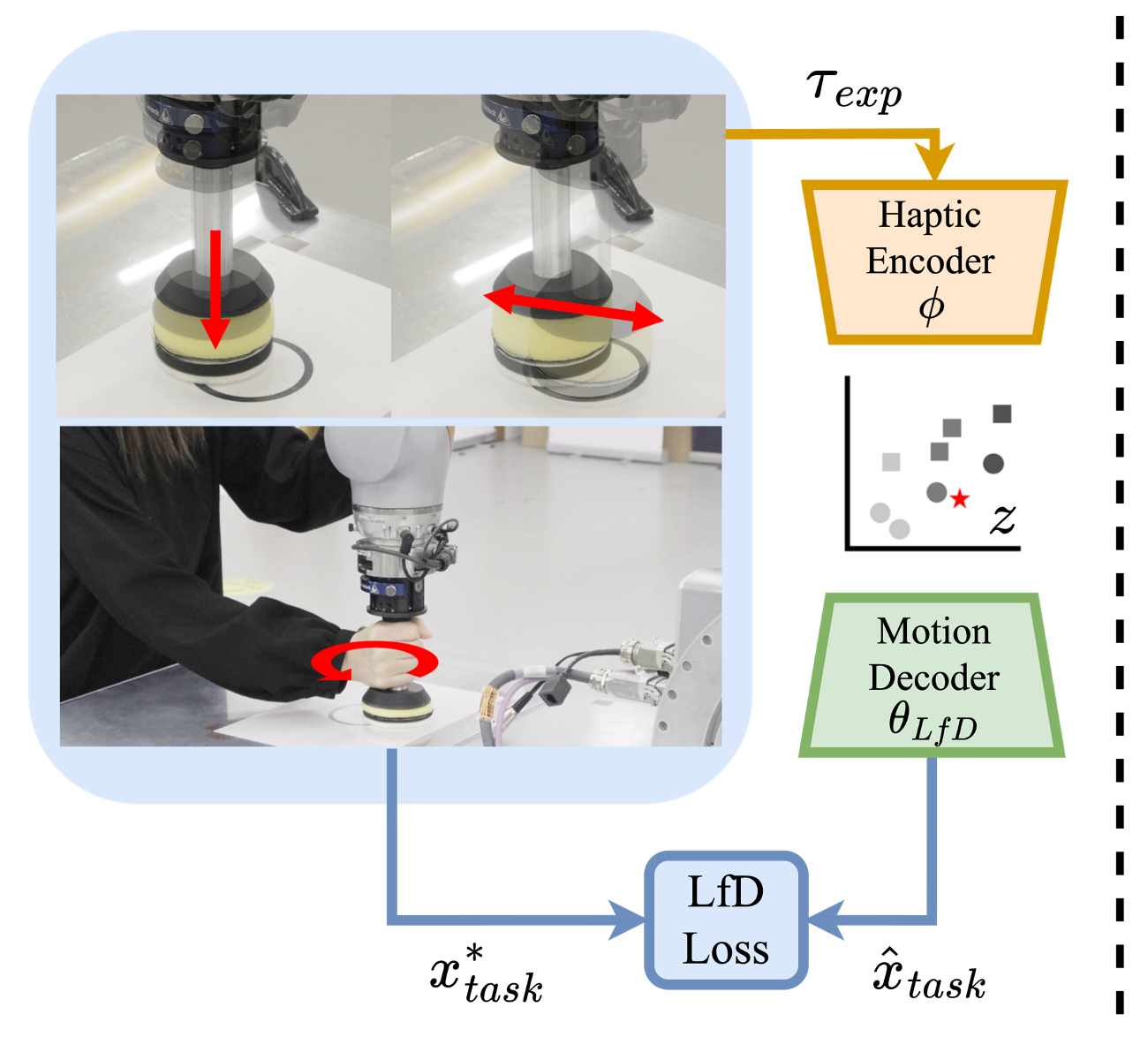}
        \subcaption{Training Step2: \\ Few-shot \gls{lfd}}
        \label{f:training_step2}
      \end{minipage} &
    \begin{minipage}[b]{0.3\hsize}
        \centering
        \includegraphics[keepaspectratio, scale=0.7]{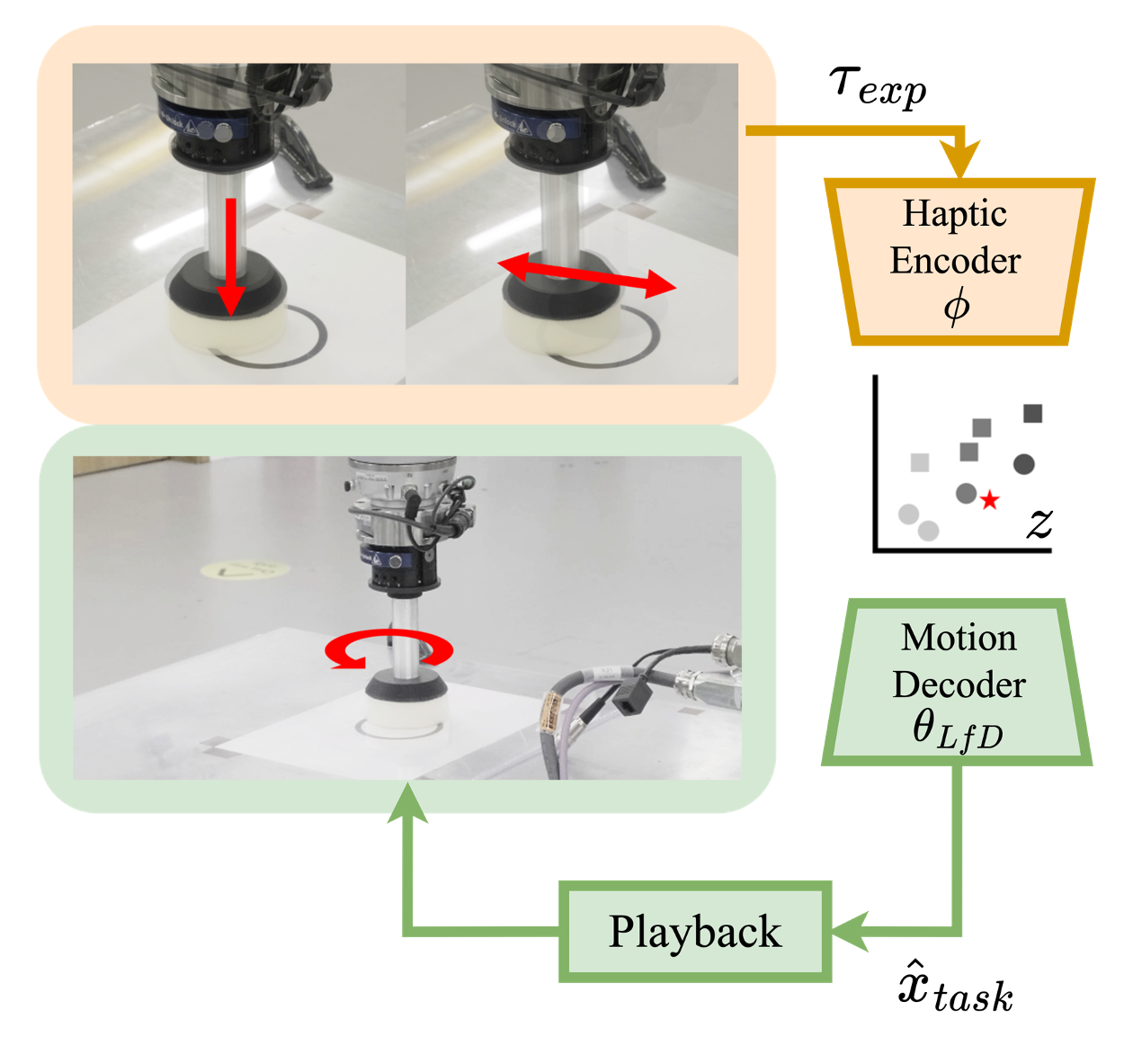}
        \subcaption{Task Execution: \\ Motion generation \& Playback}
        \label{f:testing}
      \end{minipage}
    \end{tabular}
     \caption{Proposed semi-supervised \gls{lfd} framework. 
     First, we pre-train the haptic representation encoder on unsupervised haptic data collected in simulation by minimising the \glsfirst{ssl} \eref{haptic_vae_loss}. 
     Then, we train the motion decoder to map from the haptic representation $\latent$ obtained using the pre-trained haptic encoder to the desired motion for the target task on demonstration data by minimising the \gls{lfd} loss \eref{lfd_vae_loss}. } 
     \label{f:method}
\vspace{-0.5cm}
\end{figure*}

\section{METHOD}
\label{s:method}

\paraDraft{Proposed semi-supervised LfD overview}
To utilise unsupervised haptic data, we propose a semi-supervised \gls{lfd} approach by decoupling the training of the haptic representation encoder $\hapticencoder$ where $\latent = \hapticencoder\left ( \texp \right )$ and the motion generation decoder $\lfddecoder$ where $\xtask = \lfddecoder\left ( \latent \right )$. 
In contrast to the approaches that require the joint training of the encoder $\hapticencoder$ and the decoder $\lfddecoder$, the proposed decoupling enables training a part of the model, the haptic representation encoder $\hapticencoder$, using unsupervised haptic data. 
We expect the latent representation $\latent$ to implicitly encode object characteristics. 
Note that the dimension of the latent space $\latent$ is a hyperparameter and needs to be sufficiently large to be expressive yet small enough to prevent overfitting and to be computationally efficient.
We explain the specific choice of the neural network hyperparameters for this work in \sref{model_training}. 

The proposed semi-supervised \gls{lfd} framework consists of two steps: pre-training of the haptic representation encoder $\hapticencoder$ using a large amount of unsupervised haptic data and a few-shot \gls{lfd} of the motion generation decoder $\lfddecoder$ using demonstration data, as illustrated in \fref{method}. 

\subsection{Pre-training haptic encoder}
\label{s:step1_pre-training}
\paraDraft{Step1: Pre-training haptic encoder-decoder}
In the first step, we train a haptic representation encoder-decoder model. 
The objective of the model is to capture the underlying distribution of object properties in its latent space, rather than generating the desired motion for the target task, as shown in \fref{training_step1}. 
We adopt \gls{vae}~\cite{kingma2013_arxiv_vae}, a widely used approach for self-supervised learning of perceptual representation models, to train this haptic representation encoder-decoder model on unsupervised haptic data $D_{offline}=\left \{ \texp_1, \dotsc,\texp_M \right \}$.  
To achieve this, we reconstruct a force trajectory $\texp$ obtained through pre-defined exploratory actions and minimising a \gls{ssl} 

\begin{equation}
\begin{split}
\hat\hapticdecoder, \hat\hapticencoder = \arg\min_{\hapticdecoder,\hapticencoder} L\left ( \texp \right )=\beta D_{KL}\left ( q_{\hapticencoder}\left ( z\mid \texp \right )\parallel p_{\hapticencoder}\left ( \latent \right ) \right ) + \\ E_{MSE}\left ( \texp, \texprec \right ). 
\end{split}
\label{e:haptic_vae_loss}
\end{equation}
The loss function consists of two terms: the \gls{kl} divergence $D_{KL}$ between the approximate posterior and prior distribution of the latent variable $z$, and the reconstruction loss, which is the Mean Squared Error $E_{MSE}$ between the input force trajectory $\texp$ and the reconstructed force trajectory $\texprec$. 
Here, $\hapticencoder$ and $\hapticdecoder$ represent the parameters of the haptic encoder and decoder, respectively, and $\beta$ is the regularisation coefficient. 
Once we train the haptic representation encoder $\hapticencoder$, we freeze its weights and use them in the next step.

\subsection{Few-shot \gls{lfd} of motion decoder}
\label{s:step2_few-shot_lfd}
\paraDraft{Step2: Few-shot LfD motion decoder}
In the second step, we train the motion generation decoder $\lfddecoder$ to generate the desired motion for the target task $\xtask$ according to the properties of the manipulated object, using the demonstration data as depicted in \fref{training_step2}. 
For each demonstration in $D_{demo}=\left \{\left ( \texp_1,\xtaskdesired_1 \right ), \dotsc, \left ( \texp_N,\xtaskdesired_N \right ) \right \}$, the haptic encoder $\hapticencoder$, which we pre-train in the first step, encodes the force trajectory $\texp$ obtained during the exploration phase into the haptic representation $\latent$ of the object.
Then, we train a mapping from the haptic representation $\latent$ to the desired motion trajectory $\xtask$ for the target task by minimising the \gls{lfd} loss

\begin{equation}
\lfddecoderlearnt = \arg\min_{\lfddecoder} L\left ( \latent, \xtask \right )=E_{MSE}\left ( \xtasklearnt, \xtaskdesired \right ) 
\label{e:lfd_vae_loss}
\end{equation}
where $\latent \sim q_{\hapticencoder}\left ( z\mid \texp \right )$. Here, $\lfddecoder$ represents the weights of the motion decoder. 
Since the objective is to generate motions for the target task that closely resemble those demonstrated for the object by a human demonstrator, we minimise the Mean Squared Error $E_{MSE}$ between the generated motion trajectory $\xtasklearnt$ and the demonstrated motion trajectory $\xtaskdesired$. 

\subsection{Task Execution}
\label{s:task_execution}
\paraDraft{Task execution procedure}
At test time, the robot first performs the same pre-defined exploratory actions employed in training on the object to be manipulated before executing the task. 
The haptic encoder $\hapticencoder$ encodes the force trajectory $\texp$ recorded during the exploration phase into a haptic representation $\latent = \hapticencoder\left ( \texp \right )$ of the object.
Subsequently, the motion decoder $\lfddecoder$ generates a motion trajectory $\xtask = \lfddecoder\left ( \latent \right )$ specific to the object for the target task. 
Finally, the robot plays back the generated motion trajectory $\xtasklearnt$ and executes the task without using real-time sensor feedback (\ie in an open-loop manner), as shown in \fref{testing}. 

\section{EXPERIMENTAL SETUP}
\label{s:experiment_setup}

\subsection{Wiping task}
\label{s:wiping_task}
\paraDraft{Choice of the task}
In our experiment, we use a wiping task as an instance of a force-based manipulation task as the desired wiping motion depends on the physical properties of the wiping tool.

\paraDraft{Wiping task description}
The task is to wipe a flat table with a sponge attached to the end-effector of the robot arm, as shown in \fref{experiment_setup}.
We prepare sponges with different physical properties (\ie 4 levels of stiffness $\times$ 3 levels of surface friction) to evaluate the ability of the \gls{lfd} model to adapt its wiping motion to the sponge properties unseen at training time.

\begin{table}
\centering
\caption{Training and test objects for stiffness interpolation, stiffness extrapolation, surface friction interpolation and surface friction extrapolation, with o: training and x: test. }
\begin{tabular}{c|c|c|c|c|c|c|c|c|c|c|c} 
\hline\hline
\multicolumn{12}{c}{Stiffness}                                                                                                                                                                                                            \\ 
\hline
\multicolumn{6}{c|}{Interpolation}                                                                                  & \multicolumn{6}{c}{Extrapolation}                                                                                   \\ 
\hline
\multicolumn{2}{c|}{\multirow{2}{*}{~ ~ ~ ~}}                                & \multicolumn{4}{c|}{Stiffness level} & \multicolumn{2}{c|}{\multirow{2}{*}{~ ~ ~ ~}}                                & \multicolumn{4}{c}{Stiffness level}  \\ 
\cline{3-6}\cline{9-12}
\multicolumn{2}{c|}{}                                                        & 1 & 2 & 3 & 4                        & \multicolumn{2}{c|}{}                                                        & 1 & 2 & 3 & 4                        \\ 
\hline
\multirow{3}{*}{\begin{tabular}[c]{@{}c@{}}Friction\\level\end{tabular}} & 1 & o & x & x & o                        & \multirow{3}{*}{\begin{tabular}[c]{@{}c@{}}Friction\\level\end{tabular}} & 1 & x & o & o & x                        \\ 
\cline{2-6}\cline{8-12}
                                                                         & 2 & o & x & x & o                        &                                                                          & 2 & x & o & o & x                        \\ 
\cline{2-6}\cline{8-12}
                                                                         & 3 & o & x & x & o                        &                                                                          & 3 & x & o & o & x                        \\ 
\hline
\multicolumn{12}{c}{Surface friction}                                                                                                                                                                                                     \\ 
\hline
\multicolumn{6}{c|}{Interpolation}                                                                                  & \multicolumn{6}{c}{Extrapolation}                                                                                   \\ 
\hline
\multicolumn{2}{c|}{\multirow{2}{*}{~ ~ ~ ~}}                                & \multicolumn{4}{c|}{Stiffness level} & \multicolumn{2}{c|}{\multirow{2}{*}{~ ~ ~ ~}}                                & \multicolumn{4}{c}{Stiffness level}  \\ 
\cline{3-6}\cline{9-12}
\multicolumn{2}{c|}{}                                                        & 1 & 2 & 3 & 4                        & \multicolumn{2}{c|}{}                                                        & 1 & 2 & 3 & 4                        \\ 
\hline
\multirow{3}{*}{\begin{tabular}[c]{@{}c@{}}Friction\\level\end{tabular}} & 1 & o & o & o & o                        & \multirow{3}{*}{\begin{tabular}[c]{@{}c@{}}Friction\\level\end{tabular}} & 1 & o & o & o & o                        \\ 
\cline{2-6}\cline{8-12}
                                                                         & 2 & x & x & x & x                        &                                                                          & 2 & o & o & o & o                        \\ 
\cline{2-6}\cline{8-12}
                                                                         & 3 & o & o & o & o                        &                                                                          & 3 & x & x & x & x                        \\
\hline
\end{tabular}
\label{t:training_data}
\vspace{-0.5cm}
\end{table}

\subsection{System design and robot setup}
\label{s:system_design}
\paraDraft{Robot and PyBullet setup}
We use a 7 \gls{dof} KUKA iiwa robotic arm with a sponge attached to its end-effector for both simulation and real robot experiments.
We mount a 6-axis force torque sensor on the wrist of the robot arm to measure force and torque along the x, y, and z axes. 
We control the robot by specifying the end-effector position when performing exploratory actions or replaying generated motions.
We provide demonstrations by kinaesthetically moving the robot arm in gravity compensation mode. 
For the simulation, we use a physics simulation framework, ROS-PyBullet~\cite{Christopher2022_arxiv_ros_pybullet_chrismower_paper, Pybullet}, with exactly the same setup as the real robot experiment. 

\subsection{Data collection}
\label{s:data_collection}
\subsubsection{Unsupervised haptic data}
\label{s:unsupervised_data_collection}
\paraDraft{Exploration}
In the exploration phase, the robot sequentially performs two pre-defined exploratory actions: pressing (\ie moving the end-effector in the normal direction at a speed of \SI[per-mode=symbol]{0.01}{\meter\per\second} for 2 seconds) and lateral motion (\ie moving the end-effector horizontally at speeds of \SI[per-mode=symbol]{0.05}{\meter\per\second} and \SI[per-mode=symbol]{-0.05}{\meter\per\second} for 1 second each). 
We record force and torque data along the x, y, and z directions at a frequency of 100 Hz during the exploration phase to obtain $\texp\in\mathbb{R}^{2400}$ (200 time steps $\times$ 2 exploratory actions $\times$ 6 (force and torque) in (x,y,z) axis). 

\paraDraft{Data collection in simulation}
We collect 1000 unsupervised haptic data in simulation by performing the exploratory actions on simulated objects with varying contact stiffness $k\in \left [ 80,1000 \right ]$ \si[per-mode=symbol]{\newton\per\meter}, lateral friction coefficient $\mu_{lateral}\in \left [ 0.2,8.0 \right ]$ and spinning friction coefficient $\mu_{spin}\in \left [ 0.0,4.0 \right ]$. 
Note that a robot can obtain unsupervised haptic data either in simulation or on a physical robotic setup by repeatedly performing the pre-defined exploratory actions on a variety of objects.
Data informativeness increases from unsupervised simulated data to unsupervised real data and demonstration data, with a corresponding increase in the cost of data collection. 
Unsupervised real data eliminates the issue of the sim2real gap; however, it is still more expensive to collect than unsupervised simulated data.

\subsubsection{Demonstration data}
\label{s:demonstration_data_collection}
\paraDraft{Exploration and demonstration}
First, the robot performs pre-defined exploratory actions on a sponge to obtain the exploration data $\texp$, following the same procedure as in unsupervised haptic data collection. 
Subsequently, a human demonstrator kinaesthetically performs the desired wiping motion for the sponge in gravity compensation mode. 
The demonstrator wipes the table in a circular pattern, exerting as much force as possible in the normal direction to maximise cleaning effectiveness while maintaining a smooth motion. 
The amount of force applied depends on the stiffness and surface friction of the sponge, with less force desired for a stiff or rough sponge and vice versa. 
Each wiping motion lasts for 20 seconds. 
The position of the end-effector in the x, y, and z axes is recorded at a frequency of 100 Hz during the demonstration to obtain the desired wiping motion trajectory for the sponge, denoted as $\xtask\in\mathbb{R}^{6000}$ (2000 time steps $\times$ 3 (position) in (x,y,z) axis). 

\paraDraft{Demonstration data collection}
We collect demonstrations for all 12 sponges (4 stiffness levels $\times$ 3 surface friction levels $\times$ 1 trial), and left out the test cases for evaluation. 
We train the model on 4 different sets of training data ($N=6$ or $8$), to test 4 different cases of interpolation and extrapolation of stiffness and surface friction, as shown in~\tref{training_data}.

\subsection{Model training}
\label{s:model_training}
\paraDraft{Data pre-processing}
Prior to the training, we apply a Butterworth low-pass filter to all force trajectories. 
The filter parameters used are as follows: sample rate $r=70000$, cutoff frequency $f_p=\SI{1000}{\hertz}$, stopband frequency $f_s=\SI{100}{\hertz}$, maximum allowed loss in the passband $g_{pass}=\SI{3}{\decibel}$ and maximum required loss in the stopband $g_{stop}=\SI{40}{\decibel}$.
We normalise all force and motion trajectory data to $\left [ 0, 0.9 \right ]$. 

\paraDraft{Specification of haptic encoder-decoder training and parameters}
First, we pre-train the haptic encoder, as described in \sref{step1_pre-training}, using 1000 simulated unsupervised haptic data. 
The haptic encoder-decoder model consists of 1 fully connected encoder layer, 1 sampling step, and 1 fully connected decoder layer with \gls{relu} as the activation function. 
To prevent overfitting, we apply a dropout rate of 0.1 to the decoder. 
We train the haptic encoder-decoder model for 200 epochs at a learning rate of 0.0001, using a loss function \eref{haptic_vae_loss} with $\beta=0.06$. 
We set the latent space dimension to be $\latent\in\mathbb{R}^{5}$ to implicitly represent the stiffness, surface friction, and other non-intuitive physical properties of the sponges. 
We use the Adam optimizer~\cite{kingma2014_arxiv_adam} to update the weights of the neural network.

\paraDraft{Specification of motion decoder training and parameters}
Then, we train the motion decoder as described in \sref{step2_few-shot_lfd} using demonstration data produced by a human demonstrator on the real robot. 
The motion decoder consists of 1 fully connected layer with a dropout rate of 0.1. 
We train the motion decoder for 10000 epochs at a learning rate of 0.001 using a loss function \eref{lfd_vae_loss}. 

\section{RESULTS AND DISCUSSION}
\label{s:result_and_discussion}

\begin{figure}[t]
\centering
\includegraphics[keepaspectratio, scale=0.53]{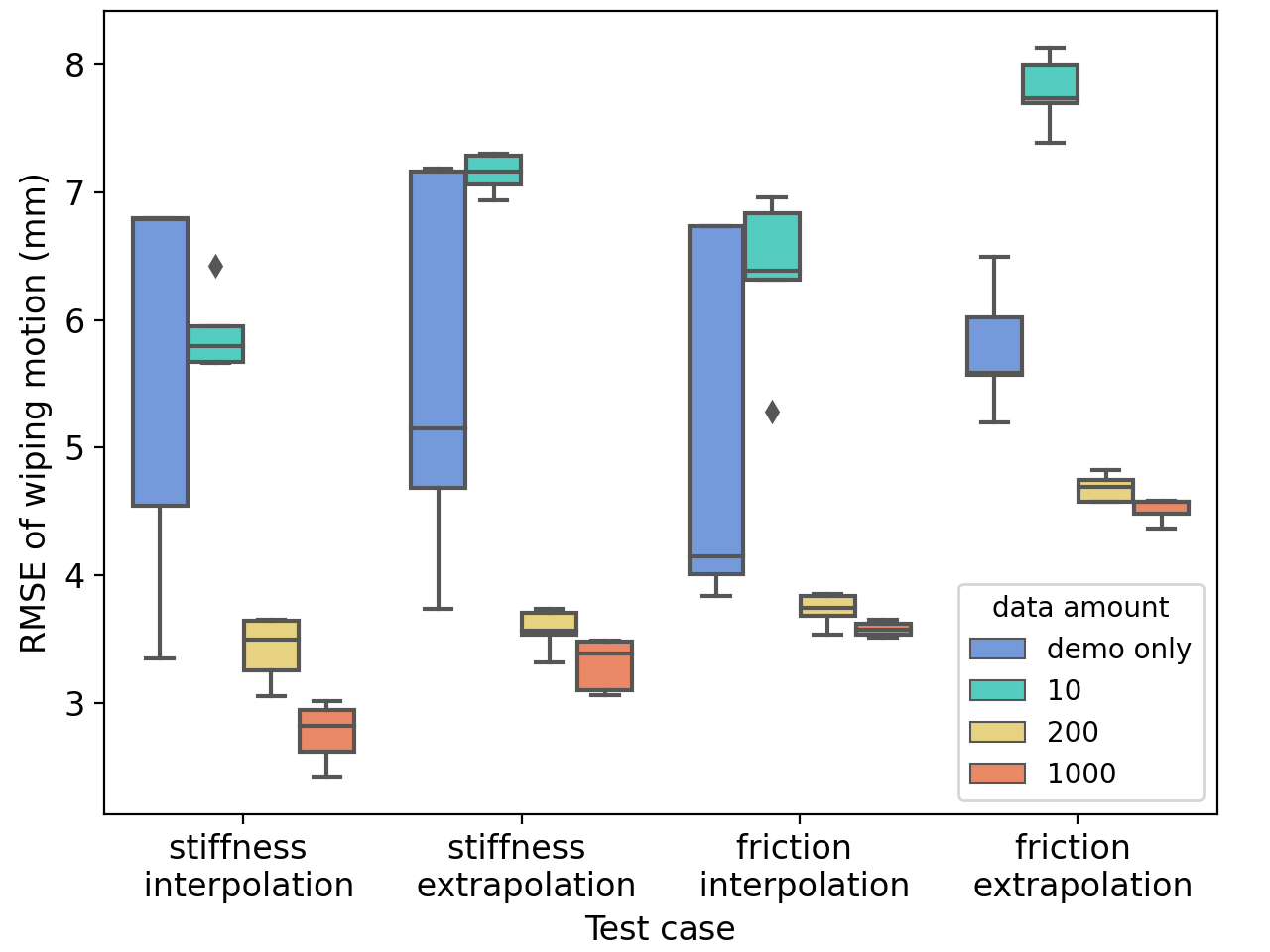}
\caption{
\glsentryshort{rmse} between the wiping motion trajectories demonstrated and generated by the \gls{lfd} model with pre-training (\ie proposed) and without pre-training (\ie baseline shown in blue).
}
\setlength\intextsep{0pt}
\label{f:datasize}
\vspace{-0.2cm}
\end{figure}


\begin{table}
\begin{adjustbox}{width=0.5\textwidth}
\centering
\begin{tabular}{ccccc} 
\hline\hline
\multirow{2}{*}{}                                                    & \multicolumn{2}{c}{Stiffness} & \multicolumn{2}{c}{Surface
  Friction}  \\ 
\cline{2-5}
                                                                     & Interpolation & Extrapolation & Interpolation & Extrapolation           \\ 
\hline
\begin{tabular}[c]{@{}c@{}}Demo only\\ (Baseline)\end{tabular}       & 12.5 ± 5.1    & 25.1 ± 22.9    & 9.0 ± 0.6     & 13.2 ± 5.0              \\
\begin{tabular}[c]{@{}c@{}}Pre-trained sim\\ (Proposed)\end{tabular} & 5.6 ± 2.2     & 6.4~ ± 3.6    & 5.4 ± 0.4     & 4.4 ± 1.7               \\
\hline
\end{tabular}
\end{adjustbox}
\caption{\glsentryshort{rmse} of force profile (N) when playing back the demonstrated and generated wiping motion for unseen objects.
}
\label{t:rmse_force}
\vspace{-0.5cm}
\end{table}

\begin{figure}[t]
\centering
\includegraphics[keepaspectratio, scale=0.43]{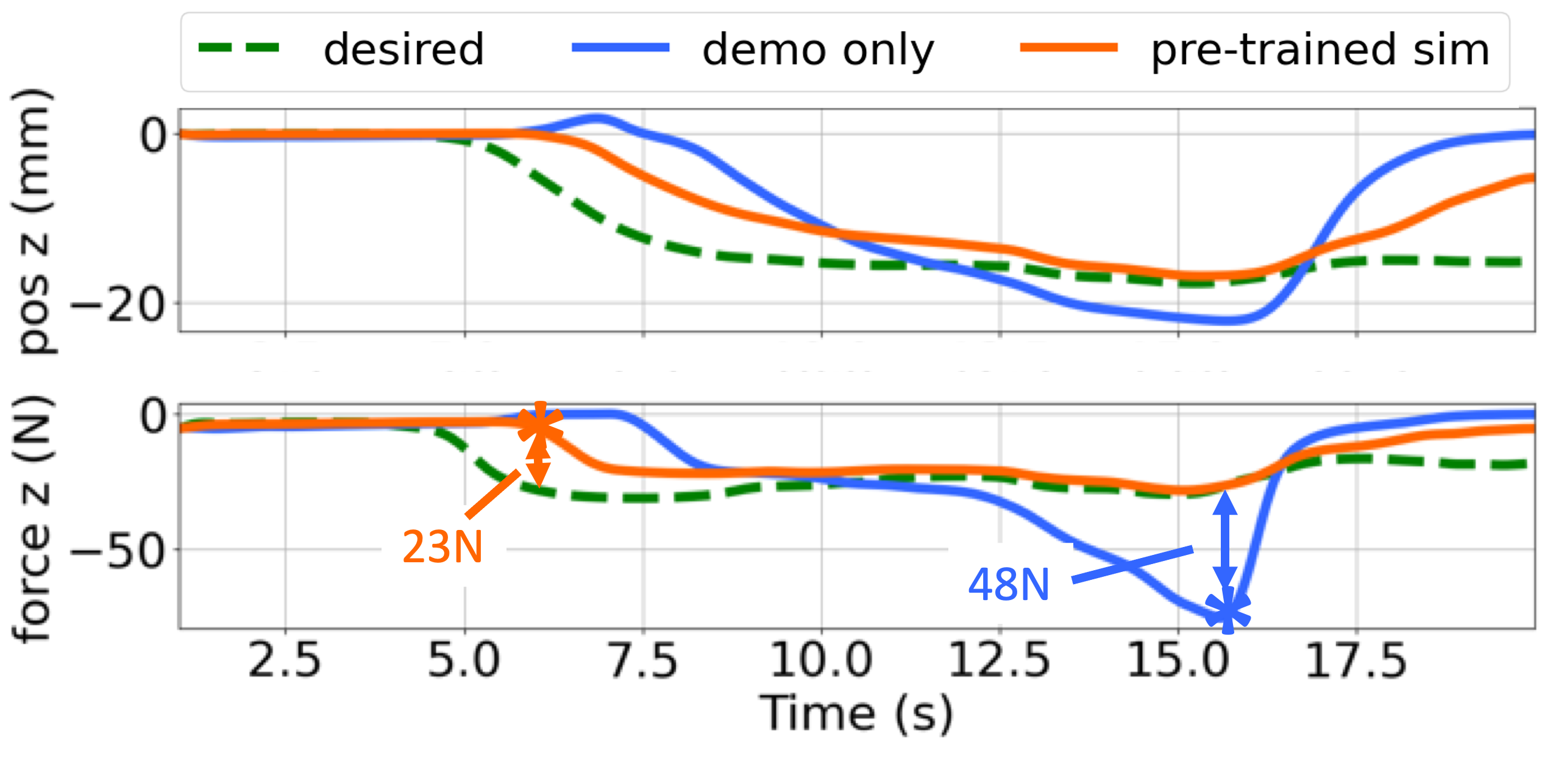}
\caption{
Plots of the wiping motion and force profile in the normal direction when playing back the wiping motion trajectory demonstrated and generated by the \gls{lfd} model with pre-training (\ie proposed) and without pre-training (\ie baseline) of the haptic encoder. }
\setlength\intextsep{0pt}
\label{f:plot_wiping_motion_and_force}
\vspace{-0.5cm}
\end{figure}

\begin{figure*}[t]
\captionsetup[subfigure]{justification=centering}
    \begin{tabular}{cccc}
      \begin{minipage}[b]{0.21\hsize}
        \centering
        \includegraphics[keepaspectratio, scale=0.38]{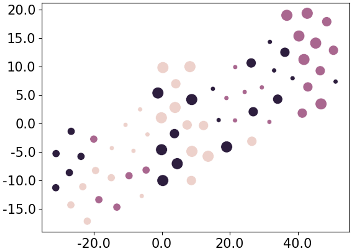}
        \subcaption{Baseline: demo only \\ $N=8$}
        \label{f:pca_real_no_pretraining}
      \end{minipage} &
      \begin{minipage}[b]{0.21\hsize}
        \centering
        \includegraphics[keepaspectratio, scale=0.39]{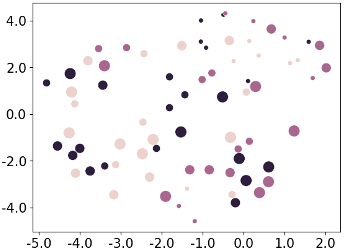}
        \subcaption{Proposed: pre-trained \\ $M=10$}
        \label{f:pca_real_pretraining10}
      \end{minipage} &
    \begin{minipage}[b]{0.21\hsize}
        \centering
        \includegraphics[keepaspectratio, scale=0.39]{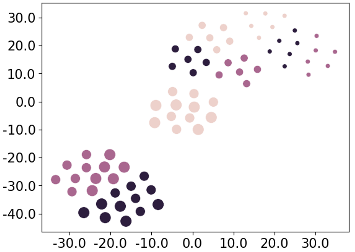}
        \subcaption{Proposed: pre-trained \\ $M=200$}
        \label{f:pca_real_pretraining200}
      \end{minipage} &
      \begin{minipage}[b]{0.28\hsize}
        \centering
        \includegraphics[keepaspectratio, scale=0.39]{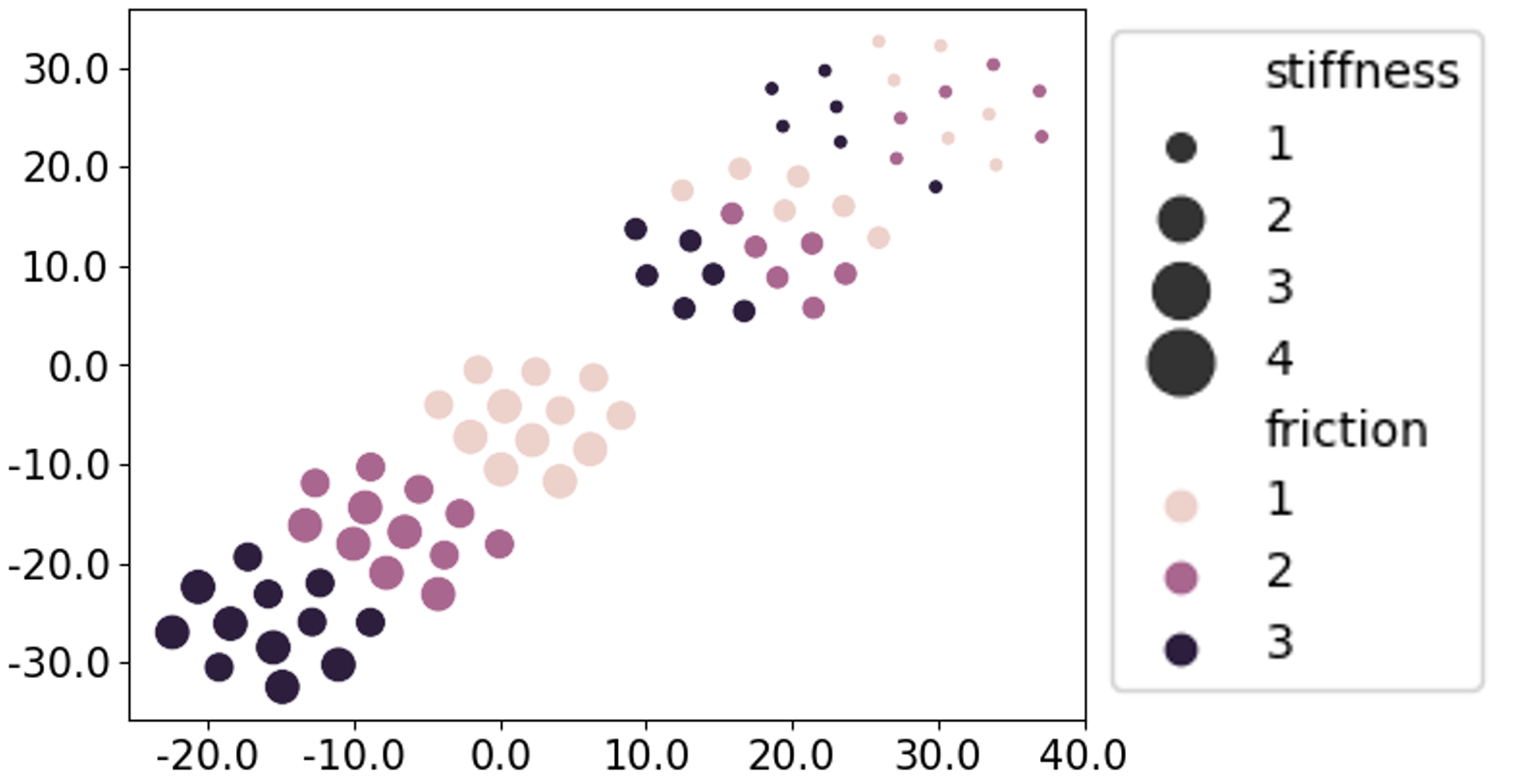}
        \subcaption{Proposed: pre-trained \\ $M=1000$}
        \label{f:pca_real_pretraining1000}
      \end{minipage}
    \end{tabular}
     \caption{\gls{tsne} applied to the latent space of the haptic encoder when tested on the real objects. The baseline model is trained only using demonstration data. The proposed models are trained with different size of simulated unsupervised data. }
     \label{f:latent_pca}
\vspace{-0.2cm}
\end{figure*}

\subsection{Does pre-training haptic representation encoder improve the generation of desired behaviour for unseen objects? }
\label{s:result_motion_generation}
\paraDraft{Results on motion generation with and without pre-training}
\fref{datasize} shows the \gls{rmse} between the demonstrated and generated motions for four test cases: stiffness interpolation, stiffness extrapolation, friction interpolation, and friction extrapolation. 
We train the model five times with randomly initialised weights. 
In all four test cases, we observe a statistically significant reduction in the \gls{rmse} of the generated wiping motion trajectory when we pre-train the model using 1000 unsupervised haptic data (orange), compared to when we use only demonstration data without pre-training (blue).
It is important to note that the proposed approach achieves this improvement despite using the same number of demonstrations for training (\ie without increasing human effort). 

In addition, the comparison of models pre-trained on different sizes of unsupervised haptic data (\ie $M=10, 200$ and $1000$) shows that the \gls{rmse} between the demonstrated and generated motion trajectories decreases as the amount of unsupervised haptic data increases. 
This result highlights the importance of pre-training the haptic representation encoder on large amounts of unsupervised haptic data, enabled by the proposed decoupling of the haptic representation encoder from the motion generation decoder.

\paraDraft{Results on force profile during wiping with and without pre-training}
Next, we played back the demonstrated and generated wiping motion trajectories for unseen sponges on a real robot. 
In all four test cases, we observe a smaller \gls{rmse} of the force profiles during the wiping with pre-training, as shown in \tref{rmse_force}, reflecting a decrease in the \gls{rmse} of the generated end-effector motions as shown in \fref{datasize}.  

\fref{plot_wiping_motion_and_force} shows an example plot of end-effector motion and force profile when playing back a wiping motion on a real robot using an unseen sponge with low stiffness and high surface friction in the case of surface extrapolation. 
In this particular example, as the surface friction is high, the human demonstrator applies less force in the normal direction to maintain a smooth wiping motion compared to the sponges used for training, which have a lower friction. 
The pre-trained model exhibits similar behaviour by generating a wiping motion with less force in the normal direction. 
In contrast, the model without pre-training generates motion that exerts more force, closer to the sponges with lower friction encountered during training. 
Notably, the maximum difference in force between the demonstrated and generated motion is \SI{23}{\newton} for the pre-trained model, while the model without pre-training exhibits a maximum force difference of up to \SI{48}{\newton}. 
The ability to adapt the motion to the sponge with higher friction, which the model without pre-training fails to adapt to, showcases the advantage of the pre-trained model in manipulating unseen objects that fall outside of the demonstrated data distribution. 

\subsection{How does the haptic representation encoder trained in simulation capture the properties of real objects? }
\label{s:result_sim2real}

\paraDraft{Latent space analysis (Baseline)}
Next, we employ \gls{tsne} to visualise and analyse how the haptic encoder captures the properties of real objects in its latent space. 
We first present an example of \gls{tsne} applied to the latent space of a haptic encoder trained jointly with a motion decoder using only demonstration data (\ie baseline) to evaluate its ability to distinguish the properties of unseen real sponges in the case of stiffness extrapolation.
The visualisation in \fref{pca_real_no_pretraining} shows a lack of clear organisation in its latent space with respect to the stiffness and surface friction of the sponges, as 6 demonstration data (2 stiffness levels $\times$ 3 surface friction levels) are insufficient for learning to distinguish the physical properties of unseen objects. 

\paraDraft{Latent space analysis (Proposed)}
On the other hand, when we train the haptic encoder on 1000 simulated unsupervised haptic data and evaluated it on real sponges (\ie proposed), the haptic encoder is able to capture both the stiffness and the surface friction of the real objects in its latent space, as shown in \fref{pca_real_pretraining1000}. 
The latent space exhibits a clear organisation from compliant to stiff sponges, as indicated by the size of the markers.
Similarly, within sponges of the same stiffness, the latent space exhibits a clear organisation from sponges with smooth to rough surface friction, as indicated by the colours of the markers. 
We observe limited organisation in the latent space with respect to the surface friction of the real objects compared to the stiffness.
One possible reason for this is due to the mismatch between the simulated and real-world force observations during exploration, arising from the larger sim2real gap in the friction model compared to the stiffness model. 

Next, we apply \gls{tsne} to the latent space of the haptic encoders pre-trained on unsupervised haptic data of different sizes ($M = 10, 200$ and $1000$), as shown in \fref{pca_real_pretraining10}-\fref{pca_real_pretraining1000}, respectively. 
The results indicate that as the data size increases, the latent space becomes more organised based on the physical properties of the objects. 
There is a strong relation between how adequately the haptic representation encoder captures the physical properties of the objects and the \gls{rmse} of the generated motion. 
This relationship highlights the necessity of capturing the underlying distribution of object properties in order to generalise motions learnt from demonstrations to unseen objects. 

\begin{figure}[t]
\captionsetup[subfigure]{justification=centering}
    \begin{tabular}{cc}
      \begin{minipage}[b]{0.37\hsize}
        \centering
        \includegraphics[keepaspectratio, scale=0.37]{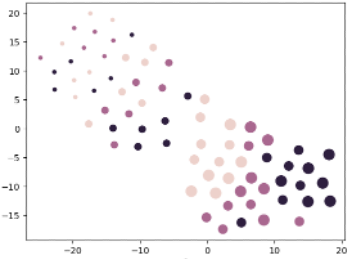}
        \subcaption{Pressing only}
        \label{f:tsne_pressingonly}
      \end{minipage} &
      \begin{minipage}[b]{0.43\hsize}
        \centering
        \includegraphics[keepaspectratio, scale=0.37]{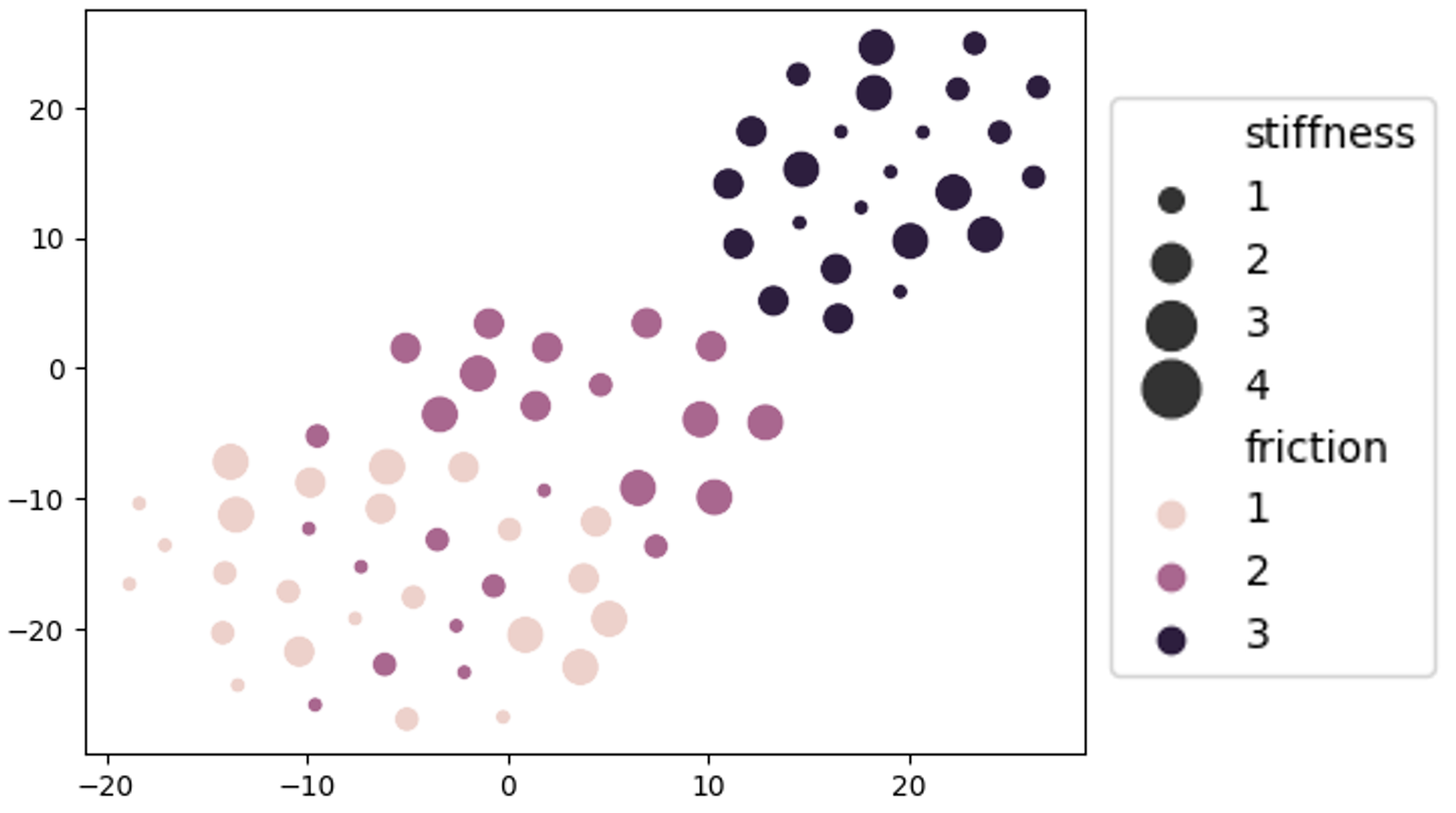}
        \subcaption{Lateral motion only}
        \label{f:tsne_lateralonly}
      \end{minipage}
    \end{tabular}
     \caption{\gls{tsne} applied to the latent space of the haptic encoder when performing only (a) pressing and (b) lateral motion. }
     \label{f:latent_tsne_eachexp}
\vspace{-0.5cm}
\end{figure}

\subsection{Does the haptic representation encoder capture the physical properties of objects? }
\label{s:result_each_exploration}
Next, we train the haptic representation encoder using only each of the exploratory actions to gain further insight into the information encoded in the latent space. 
When using force trajectories obtained from pressing only, the latent space clearly captures the stiffness of the sponges, as indicated by the marker sizes in \fref{tsne_pressingonly}, while the surface friction of the sponges remains indistinct. 
Conversely, when performing the lateral motion only, the latent space distinctly captures the surface friction of the sponges, as indicated by the marker colours in \fref{tsne_lateralonly}, while the separation with respect to the stiffness of the sponges is unclear. 
The latent space captures the properties of interest corresponding to each designed exploratory action. 
These results indicate that the haptic representation encoder encodes meaningful information (\ie the physical properties of the object) without explicitly specifying which physical properties to encode or providing access to the ground truth of the physical property parameters.

\section{CONCLUSIONS}
\label{s:conclusion}
\paraDraft{Summary of the proposed approach and results}
Our research addresses the challenge of enabling robots to learn to manipulate objects with diverse physical properties from limited human demonstrations. 
We propose a semi-supervised \gls{lfd} approach to pre-train the haptic representation encoder independently of the motion generation decoder, utilising unsupervised data without relying solely on demonstration data. 
Experimental results show that the pre-training improves the ability of the \gls{lfd} model to recognise the physical properties and generate the desired wiping motion for unseen sponges, outperforming the \gls{lfd} without pre-training. 
In addition, our analysis of the latent space of the haptic representation encoder demonstrates its capability to capture real object properties when trained on simulated unsupervised data.

\paraDraft{limitations of the proposed approach}
In this work, we carefully chose the exploratory actions considering the physical properties of the objects relevant to the downstream task. 
An interesting area for future research involves generating such exploratory behaviours by observing human exploration or by optimising information gain. 
In addition, we would like to relax the assumption of unchanged object properties during task execution and develop methods to enable online updates of object properties and adaptation of behaviours. 


\addtolength{\textheight}{-12cm}   

\printbibliography

\end{document}

%% file: ext/abb.tex
\makeglossaries

\newacronym{lfd}{LfD}{Learning from Demonstration}
\newacronym{vae}{VAE}{Variational Auto-Encoder}
\newacronym{kl}{KL}{Kullback–Leibler}
\newacronym{ssl}{SSL}{Self-Supervised Loss}
\newacronym{rmse}{RMSE}{Root Mean Squared Error}
\newacronym{pca}{PCA}{Principal Component Analysis}
\newacronym{dof}{DoF}{Degrees of Freedom}
\newacronym{tsne}{t-SNE}{t-distributed Stochastic Neighbor Embedding}
\newacronym{mae}{MAE}{masked autoencoder}
\newacronym{byol}{BYOL}{Bootstrap Your Own Latent}
\newacronym{relu}{ReLU}{Rectified Linear Unit}

%% file: ext/symbols.tex
\providecommand{\texp}{\mathnormal{\tau_{exp}}} 
\providecommand{\texprec}{\mathnormal{\bar{\tau}_{exp}}} 
\providecommand{\xtask}{\mathnormal{x_{task}}} 
\providecommand{\xtaskdesired}{\mathnormal{x^*_{task}}} 
\providecommand{\xtasklearnt}{\mathnormal{\hat{x}_{task}}} 

\providecommand{\hapticencoder}{\mathnormal{\phi}} 
\providecommand{\hapticdecoder}{\mathnormal{\theta}} 
\providecommand{\lfddecoder}{\mathnormal{\theta_{LfD}}} 
\providecommand{\lfddecoderlearnt}{\mathnormal{\hat{\theta}_{LfD}}} 

\providecommand{\latent}{\mathnormal{z}} 